\documentclass[letterpaper,10pt,journal,twoside]{IEEEtran}

\usepackage{amsmath,amsfonts}
\usepackage{algorithmic}
\usepackage{array}
\usepackage[caption=false,font=normalsize,labelfont=sf,textfont=sf]{subfig}
\usepackage{textcomp}
\usepackage{stfloats}
\usepackage{url}
\usepackage{verbatim}
\usepackage{graphicx}
\usepackage{balance}

\usepackage{hyperref}
\usepackage[capitalise]{cleveref}
\usepackage{siunitx}
\usepackage{booktabs}
\usepackage{multirow}

\newcommand{\ieeemember}{{\IEEEmembership{Member, IEEE}}}
\newcommand{\ieeegraduatestudentmember}{{\IEEEmembership{Graduate Student Member, IEEE}}}

\newcommand{\eg}{\textit{e.g.}}
\newcommand{\ie}{\textit{i.e.}}

\begin{document}

%%%%%%%%%%%%%%%%%%%%%%%%%%%%%%%%%%%%%%%%%%%%%%%%%%%%%%%%%%%%%%%%%%%%%%%%%%%%%%%%%%%%%%%%%%%%%%%%%%%%
\title{Learning Agile Swimming: An End-to-End Approach without CPGs}

%%%%%%%%%%%%%%%%%%%%%%%%%%%%%%%%%%%%%%%%%%%%%%%%%%%%%%%%%%%%%%%%%%%%%%%%%%%%%%%%%%%%%%%%%%%%%%%%%%%%
\author{Xiaozhu Lin, \ieeegraduatestudentmember, Xiaopei Liu, and Yang Wang, \ieeemember%
\thanks{Manuscript received August 19, 2024; revised October 26, 2024; accepted December 28, 2024.}% <-- only final version
\thanks{This paper was recommended for publication by Editor Jaydev P. Desai upon evaluation of the Associate Editor and Reviewers' comments.}% <-- only final version
\thanks{This work was supported by the National Natural Science Foundation of China under Grant 62072310. \textit{(Corresponding author: Yang Wang.)}}%
\thanks{Xiaozhu Lin, Xiaopei Liu, and Yang Wang are with the School of Information Science and Technology, ShanghaiTech University, Shanghai, China. \textit{(E-mail: \{linxzh, liuxp, wangyang4\}@shanghaitech.edu.cn)}}%
\thanks{Digital Object Identifier (DOI): see top of this page.}% <-- only final version
}

%%%%%%%%%%%%%%%%%%%%%%%%%%%%%%%%%%%%%%%%%%%%%%%%%%%%%%%%%%%%%%%%%%%%%%%%%%%%%%%%%%%%%%%%%%%%%%%%%%%%
\markboth{IEEE Robotics and Automation Letters, Preprint Version, Accepted December, 2024}{Lin \MakeLowercase{\textit{et al.}}: Learning Agile Swimming}

\maketitle

%%%%%%%%%%%%%%%%%%%%%%%%%%%%%%%%%%%%%%%%%%%%%%%%%%%%%%%%%%%%%%%%%%%%%%%%%%%%%%%%%%%%%%%%%%%%%%%%%%%%
\begin{abstract}

The pursuit of agile and efficient underwater robots, especially bio-mimetic robotic fish, has been impeded by challenges in creating motion controllers that are able to fully exploit their hydrodynamic capabilities. This paper addresses these challenges by introducing a novel, model-free, end-to-end control framework that leverages Deep Reinforcement Learning (DRL) to enable agile and energy-efficient swimming of robotic fish. Unlike existing methods that rely on predefined trigonometric swimming patterns like Central Pattern Generators (CPG), our approach directly outputs low-level actuator commands without strong constraints, enabling the robotic fish to learn agile swimming behaviors. In addition, by integrating a high-performance Computational Fluid Dynamics (CFD) simulator with innovative sim-to-real strategies, such as normalized density calibration and servo response calibration, the proposed framework significantly mitigates the sim-to-real gap, facilitating direct transfer of control policies to real-world environments without fine-tuning. Comparative experiments demonstrate that our method achieves faster swimming speeds, smaller turn-around radii, and reduced energy consumption compared to the state-of-the-art swimming controllers. Furthermore, the proposed framework shows promise in addressing complex tasks, paving the way for more effective deployment of robotic fish in real aquatic environments.
\end{abstract}

%%%%%%%%%%%%%%%%%%%%%%%%%%%%%%%%%%%%%%%%%%%%%%%%%%%%%%%%%%%%%%%%%%%%%%%%%%%%%%%%%%%%%%%%%%%%%%%%%%%%

%%%%%%%%%%%%%%%%%%%%%%%%%%%%%%%%%%%%%%%%%%%%%%%%%%%%%%%%%%%%%%%%%%%%%%%%%%%%%%%%%%%%%%%%%%%%%%%%%%%%
\section{Introduction}\label{sec:intro}

\IEEEPARstart{T}{he} pursuit of agile and efficient underwater robots has garnered significant attention \cite{katzschmann2018exploration,zhong2021tunable,sun2022recent} in recent years, as driven by the applications ranging from environmental monitoring \cite{sun2022recent} to underwater exploration \cite{katzschmann2018exploration}. Among the various Autonomous Underwater Vehicles (AUVs), robotic fish, thanks to their bio-mimetic appearance and structure, has emerged as a promising solution to numerous challenges that conventional AUVs struggled to address \cite{raj2016fish}, for instance, approaching aquatic life without disturbing them or their natural environment \cite{katzschmann2018exploration}. Additional advantages of fish-like AUVs include low noise emissions \cite{zhang2022simulation}, and more importantly, the potential for high efficiency and maneuverability due to their unique propulsion mechanisms \cite{raj2016fish}. 
However, despite the recent development of numerous robotic fish prototypes \cite{zhang2022simulation,yan2020efficient,mamakoukas2021derivative}, their widespread deployment across various underwater applications has yet not been realized as anticipated by the research community. This unfortunate shortfall primarily arises from the lack of a simple and efficient approach to developing the motion controllers that fully harness the capabilities of robotic fish to achieve agile swimming. 

Addressing the control challenges of robotic systems often begins with dynamic modeling. However, the complex fluid-structure interactions during swimming introduce significant uncertainty in modeling its dynamics. For the body and/or caudal fin (BCF) swimming mode-based fish-like robot considered in this work, obtaining an accurate dynamic model is quite difficult and expensive \cite{yu2016data,wang2015averaging}. This suggests that designing a model-based controller for robotic fish requires more expertise for fish-like robots. To circumvent these challenges on modeling, data-driven methods have been used for the control of robotic fish \cite{mamakoukas2021derivative,castano2020control}, which, however, have not truly demonstrated agile motions. On the other hand, many researchers in robotic fish opt for a bio-inspired, low-level control module known as the Central Pattern Generator (CPG) to mimic the locomotion of a real fish. The CPG generates coordinated rhythmic signals using periodic functions (such as trigonometric functions) without requiring high-level commands or external feedback input. While CPG is model-free, finding appropriate input parameters like swing frequency, amplitude, and offset angle for a CPG-based controller is nontrivial \cite{tian2020cfd}. Furthermore, strictly adhering to the predefined trigonometric swimming patterns fails to capture the complex motion patterns observed in natural fish swimming \cite{li2019bottom}, ultimately hindering the learning of agile motion. To simplify controller design,  and more importantly, to maximize the potential of bio-mimetic robotic fish that better utilizes hydrodynamic properties, it is crucial to release these robots from fixed parametric patterns, allowing them to make more flexible motion decisions at each step.

Among various model-free end-to-end approaches, Deep Reinforcement Learning (DRL) \cite{hwangbo2019learning,kaufmann2023champion,song2023reaching} has shown promising results in learning tasks involving continuous state and action spaces, which are relevant to robotics. However, in the context of robotic fish, DRL is still largely limited to higher-level decision-making \cite{zhang2022simulation,yan2020efficient}, such as generating input parameters for CPG, rather than directly controlling low-level actuator commands. Although end-to-end training strategies for swimming robots do exist \cite{li2021deep}\cite{wang2022learn}, they have not successfully produced agile swimming policies, leading to a decrease of enthusiasm for pursuing end-to-end control of robotic fish. We believe that these outcomes are primarily due to several challenges. First, designing an appropriate action and state space for robotic fish, which are complex fluid-coupled systems, is particularly difficult. Second, the absence of a fast and reliable virtual environment that can accurately capture the intricate interactions between the fluid and the robotic fish while allowing for extensive exploration of actions across various environmental states is a significant limitation. Third, achieving good consistency between simulated and real-world performance remains challenging. This sim-to-real gap caused by model mismatch, incorrect simulation parameters, and numerical errors is widely recognized in the robotic community, especially for the locomotion performance \cite{peng2018sim,tan2018sim}. As a result, developing a simple and efficient framework that enables robotic fish to quickly learn agile and energy efficient swimming in an end-to-end manner remains a great challenge.

\begin{figure*}[t]
    \centering
    \includegraphics[width=0.99\linewidth]{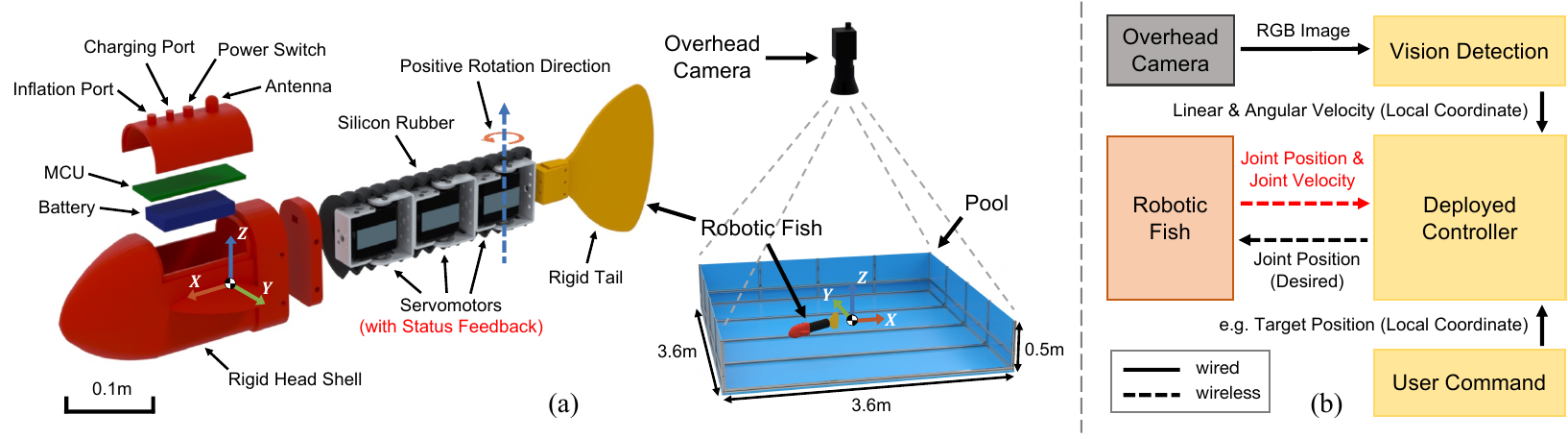}
    \vspace{-10pt}
    \caption{Overview of the experimental platform. (a) Schematic of interior details of the prototype of robotic fish and the entire experimental platform. (b) Diagram of the data processing procedure for the experiment. The modules in yellow run on a personal computer at \SI{50}{Hz} and communicate wirelessly with the robotic fish through UDP protocol.}
    \label{fig:platform}
\end{figure*}

Therefore, in order to avoid the heavy workload of designing and optimizing CPG parameters, and further release the motion agility potential of robotic fish from a single parameter CPG. This work provides a novel end-to-end RL-based agile policies generation framework for robotic fish, which utilizes the CFD fluid simulation engine previously developed by our team to simulate the dynamics of robotic fish. The shape representation of robotic fish (\ie, the current joint information of the robotic fish) is integrated into the state of the robotic fish to directly obtain the expected joint position rather than the parameters of the bottom-level periodic module. This greatly simplifies the generation steps of robotic fish motion strategies and further enhances the agility of robotic fish. Meanwhile, this article also proposes two key sim-to-real techniques, normalized density matching and servo response matching, which successfully transfer the zero-shot sim-to-real trained end-to-end multi-modal agile swimming policies. The comprehensive experimental results indicate that the proposed framework can generate end-to-end agile swimming strategies that are highly competitive compared to existing state-of-the-art methods.

Overall, the distinctive features of the proposed framework are threefold: 1) \textbf{Convenient.} The proposed framework does not rely on the dynamic model and the bottom-level periodic module, simplifying agile swimming policy design significantly. 2) \textbf{Agile.} The generated policy using proposed framework can be directly transferred to real-world to achieve agile swimming (\eg, a challenging 180 degree turn in place) without any fine-tuning. 3) \textbf{Extendible.} The proposed framework can be easily extended to other fluid interaction robots (such as manta rays-like robots or flapping wing robots) without significantly increasing workload.

%%%%%%%%%%%%%%%%%%%%%%%%%%%%%%%%%%%%%%%%%%%%%%%%%%%%%%%%%%%%%%%%%%%%%%%%%%%%%%%%
\section{Robotic Fish and Experimental Platform}\label{sec:platform}
In this section, we presents the experimental platform designed to evaluate controllers in real-world scenarios, as shown in \cref{fig:platform}. The platform includes a robotic fish prototype with joint status feedback, a large pool, and an overhead camera for global vision detection. The deployed controller runs on a personal computer, wirelessly sending commands to the robotic fish at \SI{50}{Hz}.

\subsection{Robotic Fish}
The robotic fish prototype mimics a typical carangiform fish with Body and/or Caudal Fin (BCF) propulsion mode, as depicted in \cref{fig:platform}(a). Its dimensions are \SI{0.52}{m} $\times$ \SI{0.13}{m} $\times$ \SI{0.12}{m}, weighing \SI{1.1}{kg}. The prototype is comprised of three main components: (i) a rigid head housing an MCU board with the ESP32-S3 module and an \SI{800}{mAh} \SI{7.4}{V} lithium battery; (ii) a deformable body consisting of three joints connected by aluminum components and covered with rubber skin; and (iii) a rigid tail. The head and tail are 3D-printed from ABS plastic and coated with epoxy resin for waterproofing. Each joint is driven by a Hiwonder LX-824 servomotor with a maximum torque of \SI{17}{kg\cdot cm}, a no-load speed of \SI{0.2}{s} / \SI{60}{\degree}. The density of the prototype is adjusted slightly lower than water to focus on its planar motion. A key feature distinguishing our prototype from others is its capability to provide real-time joint position feedback, enhancing state representations for the controller.

\subsection{Experimental Platform}
The real-world experiments are conducted on our developed platform, which consists of a pool, an overhead camera, and a computer, as depicted in \cref{fig:platform}. The swimming pool with dimensions of \SI{3.6}{m} $\times$ \SI{3.6}{m} $\times$ \SI{0.5}{m}, while the water depth is set to \SI{0.4}{m} during the experiment. The robotic fish receives commands (the desired joints position) from a personal computer via WiFi module to adjust servomotors, and provides real-time feedback on the current joints position. Furthermore, the overhead camera is suspended \SI{3}{m} above the ground, used to capture the position and orientation information of the robotic fish, which is processed into proprioceptive information (surge, sway, and yaw) for control. The deployed controller operates on a personal computer using the Robot Operating System (ROS) \cite{quigley2009ros} for data exchange at \SI{50}{Hz} during experiments. It is worth noting that while global information of the robotic fish is available, only information in local coordinate is utilized for controller.

\begin{figure*}[t]
    \centering
    \includegraphics[width=0.99\linewidth]{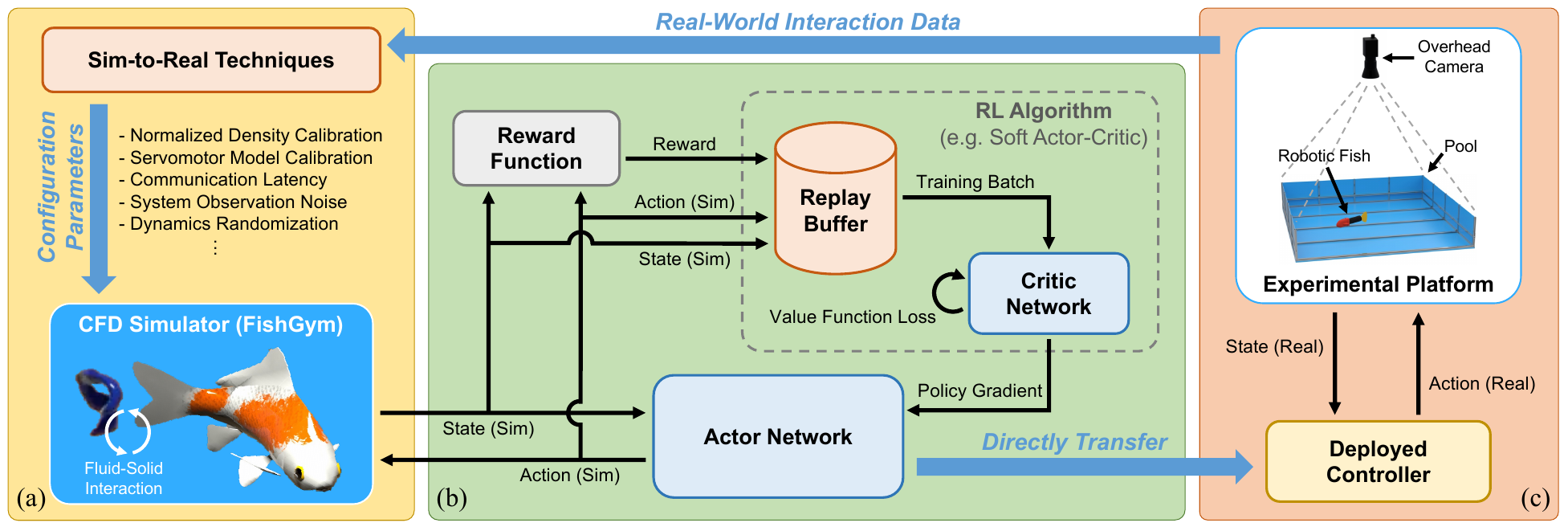}
    \vspace{-5pt}
    \caption{The proposed learning framework. (a) High-performance CFD simulator with proposed sim-to-real techniques. Only a small amount of real-world data (\eg, 30 seconds) calibration is needed to provide accurate three-dimensional fluid interaction dynamics. (b) Learning end-to-end policy from scratch. By using deep reinforcement learning in the calibrated CFD simulator, agent can learn agile swimming policy that can be directly transferred to the real-world without fine-tuning. (c) The developed robotic fish experimental platform. Used to evaluate the swimming performance of various robotic fish controllers.}
    \label{fig:framework}
\end{figure*}

%%%%%%%%%%%%%%%%%%%%%%%%%%%%%%%%%%%%%%%%%%%%%%%%%%%%%%%%%%%%%%%%%%%%%%%%%%%%%%%%
\section{Methodology}\label{sec:method}
In this section, we provide a detailed description of the proposed framework, as shown in \cref{fig:framework}. This framework aims to achieve implementation of agile swimming through end-to-end learning from scratch without any dynamic models or bottom-level periodic modules (\eg, CPG).

% we first describe the neural network controller, its state and action spaces, and its reward function. Then, we introduce the CFD simulator for policy training, and propose calibration techniques, which are the essential parts of successful zero-shot deployment to the real-world. Finally, we detail the policy training process using SAC algorithm.

\subsection{Neural Network Controller}
As one of the most common tasks for robotic fish, the position control task \cite{li2019bottom,yan2020efficient} requires an accurate approach to a given target. We formulate the this task as the standard Markov Decision Process (MDP), which is described by a tuple $(S, A, r, D, P_{sas'})$ with state space $S$, action space $A$, reward function $r$, initial state distribution $D$, and state transition probabilities $P_{sas'}$. At the start of each episode, the environment provides an initial state $s_{0}$ according to the distribution $D$. The agent selects an action $a_{0}$ using the policy function $\pi$ based on the current state $s_{0}$. Subsequently, the environment provides a reward $r_{0}$ and transitions to the next state $s_{1}$ following the transition probability $P_{sas'}(s_{t+1}|s_{t}, a_{t})$. This cycle repeats until the episode concludes. Thus, the goal of the robotic fish is to find an optimal policy \(\pi^{*}\) that maximizes the expected return of an episode as $\pi^{*} = \arg\max_{\pi} E_{\pi}[\sum_{t=0}^{T} \gamma^{t}r_t]$, where the $\gamma \in [0, 1)$ is the discount factor and $T$ is the horizon of each episode.

\subsubsection{State Space}
% what is the state & how we design the state in this work
At each timestep $t$, the RL agent receives the state $\mathbf{s}_t$ of the environment. In this work, the state consists of proprioception-related and task-related variables, denoted as $\mathbf{s}_t = [\mathbf{v}, \omega, \mathbf{J}, \dot{\mathbf{J}}, \mathbf{p}]$, where $\mathbf{v} \in \mathbb{R}^{2}$ and $\omega \in \mathbb{R}$ are the linear and angular velocities of the robotic fish in the body coordinate frame, respectively. $\mathbf{J} = [\theta_1, \theta_2, \theta_3] \in \mathbb{R}^{3}$ and $\dot{\mathbf{J}}=[\dot{\theta}_1, \dot{\theta}_2, \dot{\theta}_3] \in \mathbb{R}^{3}$ are the angular position and velocity of the three joints, respectively. $\mathbf{p} = [p_x, p_y] \in \mathbb{R}^{2}$ represents the target position in the body frame of the robotic fish.

% the advantages of our state definition
Unlike mainstream works on robotic fish \cite{zhang2022simulation,yan2020efficient}, we particularly incorporate joint-related information $\mathbf{J}$ and $\dot{\mathbf{J}}$ into our state $\mathbf{s}_t$. This choice arises from two key motivations: First, the deformation capability of robotic fish implies that even at consistent linear and angular velocities, different shapes can produce varying lift and drag forces, thereby affecting its entire coupled dynamics. Second, due to fluid resistance and restrictions in actuator response, $\mathbf{J}$ may not immediately achieve the desired joints position $\mathbf{J}^\text{des}$. Therefore, including $\mathbf{J}$ and $\dot{\mathbf{J}}$ in the state not only captures the shape representation of the robotic fish, expanding the state space to maintain the Markov property, but also enables adaptive policy adjustments to achieve joint closed-loop at the controller level.

\subsubsection{Action Space}
% what is the action & how we design the action in this work
According to the observed state $\mathbf{s}_t$, the neural network controller has to output an action $\mathbf{a}_t$ (\ie, the control command), which consists of the desired angular position of each joint, denoted as $\mathbf{a}_t = [\mathbf{J}^\text{des}] \in \mathbb{R}^{3}$. These commands are then directly applied to the joints of the robotic fish without relying on any bottom-level periodic modules.

% the advantages of our action definition
This design of the action space offers two main advantages: First, this end-to-end action space simplifies the design of the entire controller by eliminating the need to build and adjust bottom-level periodic modules, avoiding the additional expert knowledge and multifarious debugging time required. Second, it provides a sufficiently large action space for RL algorithms to explore, which can fully utilize the agility of robotic fish compared to using the several parameters of the overly simplified bottom-level modules as a limited action space.

% the challenges of our action (put into the simulator part)
Although the design of the action space is straightforward, existing literature has struggled to learn an efficient swimming policy using this action space. The primary challenge has been the lack of a sufficiently accurate and efficient virtual environment capable of capturing the transient dynamics of robotic fish for training reinforcement learning methods---a challenge this work aims to overcome.

\subsubsection{Reward Function}
The optimization goal of the swimming policy is to complete position control task in a smooth, agile, and energy-efficient manner. Therefore, the reward function is designed to encompass various optimization objectives, as follows:
\begin{equation}
\begin{aligned}
    r_t &= r^\text{appr}_t - r^\text{ener}_t + r^\text{term}_t, \\
    r^\text{appr}_t &= \lambda_1 (d_{t-1} - d_t), \\
    r^\text{ener}_t &= \lambda_2 ||\dot{\mathbf{J}}||^{2}, \\
    r^\text{term}_t &= \begin{cases} 10, \quad \text{if $d_t < 0.05$}, \\ -10, \quad \text{if outside domain} \\ 0, \quad \text{otherwise}. \end{cases}\\
\end{aligned}
\end{equation}
where $d_t$ is the distance between the robotic fish and target position at timestep $t$, and $||\dot{\mathbf{J}}||$ is the norm of the velocities of all joints. The approach reward $r^\text{appr}_t$ encourages the robotic fish to swim towards the target position, while the energy penalty $r^\text{ener}_t$ penalizes the sudden changes in joint position, the terminal reward $r^\text{term}_t$ encourages the robotic fish to approach the target instead of reaching the outside of simulated domain. The hyper-parameters $\lambda_1 = 10$ and $\lambda_2 = 0.001$ are selected empirically in this work. If $\lambda_2$ further increased to $\lambda_2 = 0.2$, the trained policy will demonstrate more tendency to glide to the target instead of performing swinging tail, which reduces energy consumption but increases the swimming time.

\subsection{CFD Simulator}\label{sec:simulator}
% why we need simulation
The RL algorithm requires a large amount of interaction data between the agent and the surrounding environment to learn complex policy from scratch, which makes direct learning from the real-world impractical, especially considering factors such as battery fluctuations, extensive manual resets, and even structural damage. To efficiently train a reliable policy within a reasonable time and transfer it to the real-world, we needed a simulation platform that is both fast and accurate. 

% why we need fast CFD simulation & FishGym brief description
One of the biggest challenges for the robotic fish is the complex dynamics of the surrounding fluid flow. To this end, we employed the three-dimensional fluid-structure interaction solver \emph{FishGym} presented in our previous work \cite{liu2022fishgym}, which leverages the highly parallel computing capabilities of the Lattice Boltzmann Method (LBM) for acceleration over GPUs. This significantly enhances the simulation efficiency, which makes this solver not only accurate but also fast, simulating for a physical time of \SI{1}{s} only takes a computing time of \SI{2}{s}, on an ordinary personal computer\footnote{All simulations (including training and evaluation) in this work were conducted on personal computers equipped with Intel Core i9-12900K CPU and NVIDIA GeForce RTX 3080TI GPU.}. Readers can refer to \cite{liu2022fishgym, lin2023exploring} for more details.

% sim-to-real challenges
Although FishGym provides a viable simulation environment, significant differences still exist between the simulated robot and its real-world counterpart in terms of appearance, linkage density, and actuator type. These differences make it challenging to directly transfer trained policies to the real world. To this end, we have explored a series of calibration techniques between the simulated and the physical robotic fish, with the following two key innovative techniques proving particularly effective in narrowing the sim-to-real gap and facilitating zero-shot transfer.

\begin{figure*}[t]
    \centering
    \includegraphics[width=0.99\linewidth]{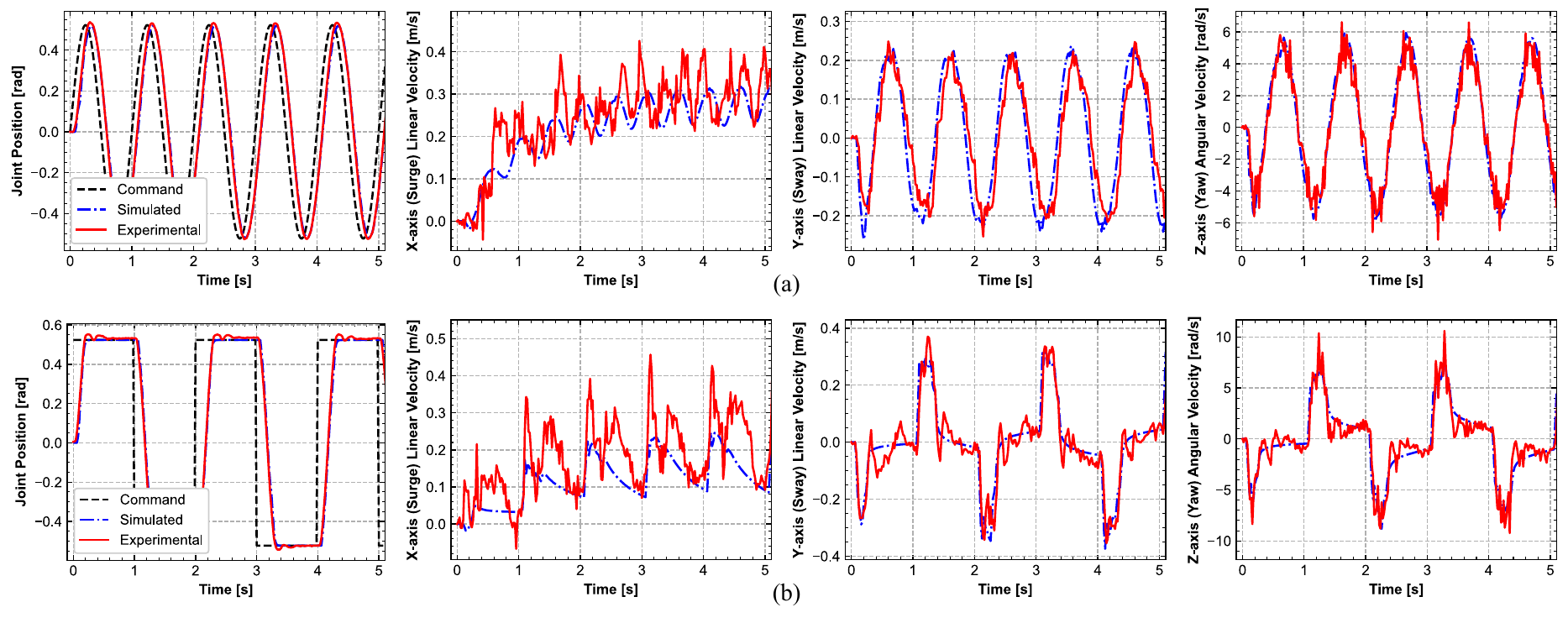}
    \vspace{-10pt}
    \caption{Validation of the calibrated CFD simulator using the proposed techniques. The the simulated results in the CFD simulator (blue) and the measured results in experimental platform (red) of the robotic fish states (Surge, Sway, and Yaw) are shown. Two desired signal (black) with different types and frequency were used for testing, (a) 1 Hz sinusoidal signal and (b) 0.5 Hz square wave signal, with amplitudes of 0.52 rad (\ie, 30 degree) for both signals. All three joints follow the same signal, and for a clearer display, only the change curve of the first joint (near the head) is shown here.}
    \label{fig:calibration}
\end{figure*}

\subsubsection{Normalized Density Calibration}\label{sec:density}
The normalized density calibration technique is used to ensure that the forces and torques brought about by joint motion can bring more similar effects. Specifically, we accurately measured the weight and length of different links in the developed robotic fish, and calculated their normalized density $\bar{\rho}_{i}$ as :
\begin{equation}
    \bar{\rho}_{i} = \frac{m_{i}}{l_{i}} \cdot \frac{\sum_{k=1}^{n}l_{k}}{\sum_{k=1}^{n}m_{k}} ,
\end{equation}
where $m_{i}$ and $l_{i}$ represent mass and length of the $i$-th link, respectively. Then, we set the corresponding links of the Koi in CFD simulation with the same normalized density as the physical robotic fish. Normalized density calibration ensures that the overall force acting on the body of the robotic fish is similar in both simulated and real environments. A mismatch in density would alter the feedback loop, creating a gap in the effectiveness of the control policy when transferred to the real-world. By highlighting normalized density calibration as a contribution, we underscore its role as a practical and reproducible strategy to improve the robustness of the underwater robot control, making the simulation-based training results more transferable.

\subsubsection{Actuator Response Calibration}\label{sec:actuator}
The joints of the robotic fish in real-world are controlled by servomotors, while the joints in FishGym are set to torque control. Therefore, we use a first-order PD controller to simulate the servomotor model $\mathbf{T} = \mathbf{k_p} * (\mathbf{J}^\text{des} - \mathbf{J}) + \mathbf{k_d} * (\mathbf{0} - \dot{\mathbf{J}})$. This model takes the action (\ie, the desired joints position $\mathbf{J}^\text{des}$) as input and outputs the execution torque $\mathbf{T}$ to FishGym to simulate subsequent fluid dynamics. By adjusting the response gain $\mathbf{k_p} \in \mathbb{R}^{3}$ and $\mathbf{k_d} \in \mathbb{R}^{3}$ of the model to achieve a same actuator response curve in the real-world.

In addition, considering the communication latency in sending control signals wirelessly to the robotic fish in real-world, we also simulated such latency by recording the history of commands  $\mathbf{J}^\text{des}_{t}$, and selecting past command instead of current ones based on the delay time $\Delta t$. That is, the used command $\mathbf{J}^\text{used}_t$ in timestep $t$ is defined as $\mathbf{J}^\text{used}_t = \mathbf{J}^\text{des}_{t - \Delta t}$.

Finally, we use two standard signals to evaluate the matching degree between the calibrated simulator and reality, as shown in \cref{fig:calibration}.

\subsection{Policy Training Details}
% using SAC algorithm
In this work, we use the Soft Actor-Critic (SAC) algorithm \cite{haarnoja2018soft} to train our policy. The SAC algorithm adds an entropy regularization term into the original AC framework, and uses temperature coefficient $\alpha$ to achieve a focus on \emph{exploration} in the early stages of training and \emph{exploitation} in the later stages, which effectively balances the exploration-exploitation trade-off and further improve sampling efficiency. This feature is very useful for CFD simulators which have a high computational cost, and can greatly reduce the number of interactions, thereby shortening the training time required. The hyper-parameters used for the SAC algorithm are listed in \cref{tab:sac}.

% randomization & noise
Although the calibration techniques described in \cref{sec:simulator} can greatly narrow the gap between simulation and reality, the uncertainty of measurement and time-varying factors such as structural wear and battery fluctuations make it difficult to determine the exact dynamics of the robotic fish. Therefore, we employ the widely recognized dynamics randomization technique \cite{peng2018sim, loquercio2019deep} to improve the robustness of trained policy to ensure zero-shot transfer. This involves randomizing key parameters that significantly impact the dynamics, uniformly sampling them within a specific range of each episode. Additionally, we add system observation noise at each step to further enhance robustness. The overall randomized parameters are shown in \cref{tab:randomization}.

\begin{table}[t]
\centering
\caption{The Soft Actor-Critic (SAC) Hyper-parameters}
\label{tab:sac}
\begin{tabular}{c|c}
    \toprule
    \textbf{Parameters}                   & \textbf{Value} \\ 
    \midrule
    discount factor ($\gamma$)            & 0.99 \\
    replay buffer size                    & $10^6$ \\
    batch size                            & 256 \\
    learning rate                         & $3 \times 10^{-4}$ \\
    % hidden layers (all networks)          & MLP[256(ReLU), 256(ReLU)] \\
    target smoothing coefficient ($\tau$) & 0.005 \\
    learning step                         & 1 \\
    entropy target                        & -3 \\
    \bottomrule
\end{tabular}
\end{table}

% training detail
At the beginning of each training episode, the robotic fish and the target point are randomly placed within the simulation domain. An episode concludes if the robotic fish moves outside the simulation domain, arrives at the target, or reaches the maximum timestep limit of $t_\text{max} = 1000$. The control interval is \SI{0.02}{s}. The structure of actor network (\ie, neural network controller) is Multi-Layer Perceptron (MLP) with [11 (ReLU), 256 (ReLU), 256 (ReLU), 3 (Tanh)], and critic network (\ie, Q network) is also MLP with [14 (ReLU), 256 (ReLU), 256 (ReLU), 1].

\begin{table}[t]
\centering
\caption{The Randomized Parameters and Their Respective Ranges}
\label{tab:randomization}
\begin{tabular}{c|c|c|c}
    \toprule
    \textbf{Parameter}      & \textbf{Distribution} & \textbf{Range}                   & \textbf{Period} \\ 
    \midrule     
    Command latency         & Uniform               & $\SI{0.068}{s} \pm \SI{0.02}{s}$ & episode \\
    Position feedback       & Gaussian              & $\pm \SI{0.05}{m}$               & step \\
    Direction feedback      & Gaussian              & $\pm \SI{3}{deg}$                & step \\
    Joint position feedback & Gaussian              & $\pm \SI{2}{deg}$                & step \\
    \bottomrule
\end{tabular}
\end{table}

%The used hyper-parameters of the SAC algorithm is shown in \cref{tab:sac}. It is worth noting that the final layer of the actor network needs to be mapped to $(-1, 1)$ using the $\tanh$ function, and then multiplied by an appropriate coefficient to re-scale to the action range. Moreover, the entropy target is defined as $-dim(A) = 3$.

\begin{figure*}[t]
    \centering
    \includegraphics[width=0.99\linewidth]{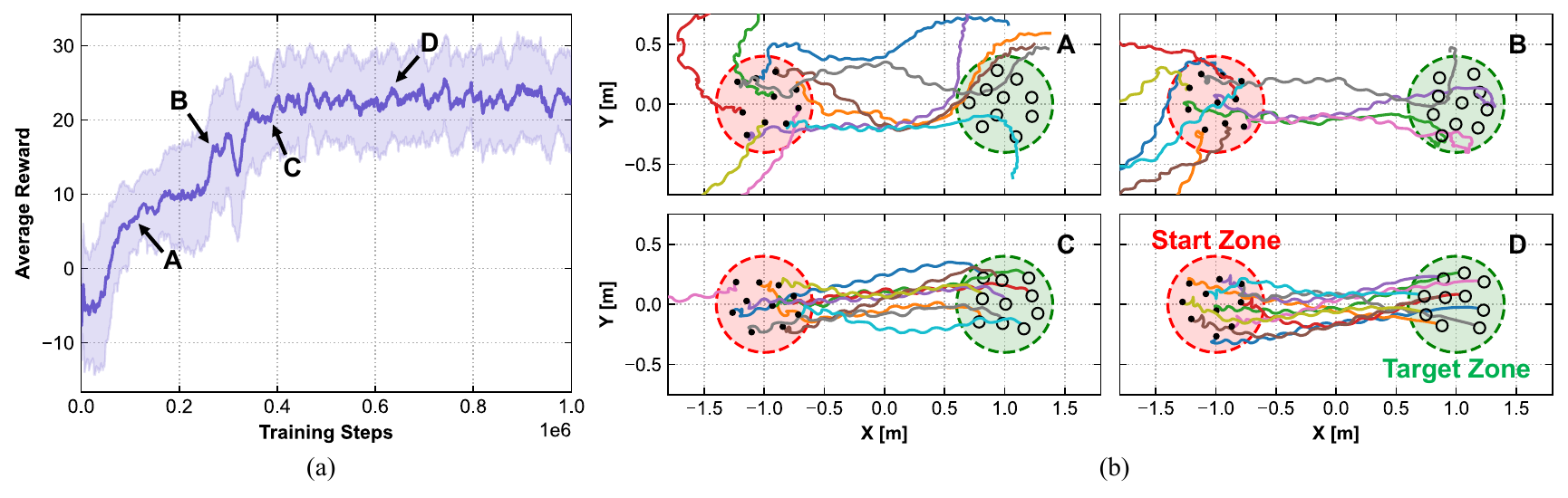}
    \vspace{-10pt}
    \caption{Evaluation of policies in different training stages. (a) Learning curves of the average reward during the training process. The training was repeated three times, with the solid line representing the mean and the shaded area representing the standard deviation range. (b) The trajectory generated by the robotic fish when completing position control at four typical learning stages: A, B, C, and D. Each evaluation is repeated 10 times, and the initial and target points are randomly sampled within the start zone (red) and target zone (green), respectively.}
    \label{fig:training}
\end{figure*}

\begin{figure*}[t]
    \centering
    \includegraphics[width=0.99\linewidth]{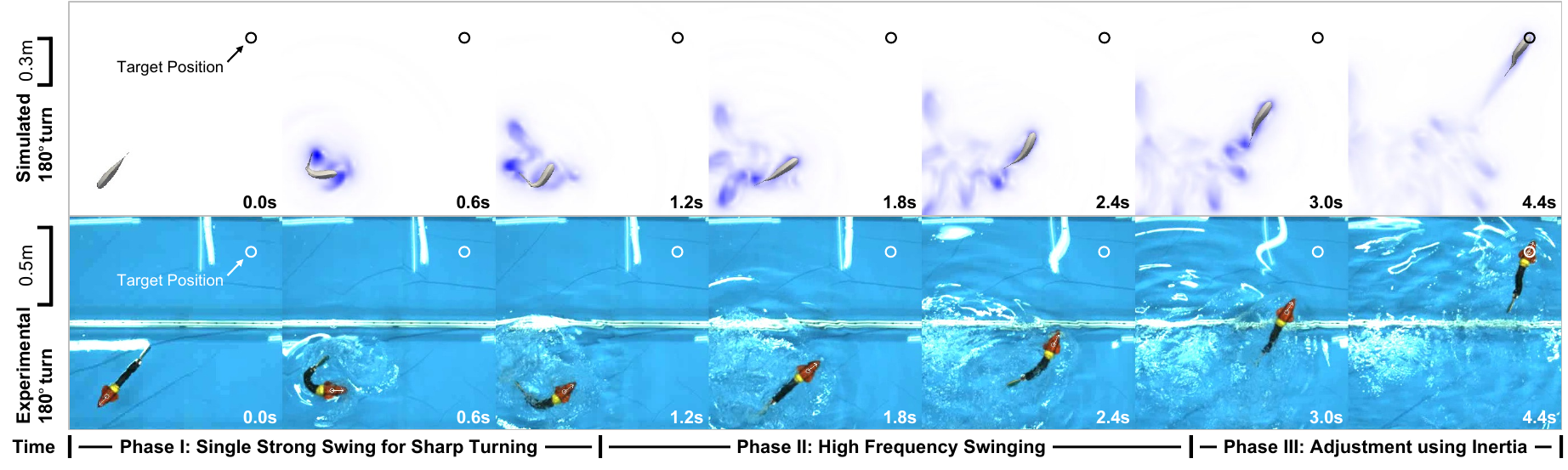}
    \vspace{-10pt}
    \caption{Screenshot sequence of the robotic fish performing a 180 degree sharp turn maneuver in both simulation and experiment. There are two points worth noting, one is the high degree of overlap between simulation and physical movement, indicating the success of sim-to-real policy transfer. The second point is that the turning maneuver of the robotic fish is very fast, requiring a small turning radius, which proves the success of end-to-end agile policy learning.}
    \label{fig:snapshot}
\end{figure*}

%%%%%%%%%%%%%%%%%%%%%%%%%%%%%%%%%%%%%%%%%%%%%%%%%%%%%%%%%%%%%%%%%%%%%%%%%%%%%%%%
\section{Experiments}\label{sec:experiments}
In this section, we comprehensively evaluated the effectiveness of the proposed end-to-end learning-based framework, to answer three extremely important questions that have never been answered by existing works: (i) Is it feasible to train an end-to-end agile swimming policy using pure CFD simulator? (ii) Is it feasible to directly transfer the trained agile swimming policy to the real world? (iii) Is the end-to-end policy more agile compared to other controllers based on bottom-level periodic module? Specifically, we first demonstrate the reward convergence curve of the learning process and the performance of the policies at several typical learning stages. Secondly, we directly transfer the policy trained in pure CFD simulation into reality, and verify the it's agility through a highly challenging task, which involves turning 180 degrees in place. Finally, we compare the trained policy with several existing state-of-the-art robotic fish control methods, and evaluate their stability during straight cruising and maneuverability during large angle turns through a pentagram waypoints tracking task.

\subsection{Training Process and Evaluations}
The whole process of learning end-to-end agile swimming policy using the proposed framework is shown in \cref{fig:training}. It is evident that as the reward function increases during the training process, the performance of the training policy in position control also improves until convergence. At first (around A stage), the robotic fish learned how to swing its tail towards the target point. Then (around B stage), the robotic fish realized the need for more precise control of its swing to avoid passing by the target point. Later (around C stage), the robotic fish successfully reached the target point but had not yet learned to use inertia to save energy during the last short distance. Finally (around D stage), the robotic fish further optimizes its strategy to achieve the optimal balance between energy and motion. All training process takes a total of 11 hours for whole $1 \times 10^{6}$ steps training from scratch. The training process almost converged after interacted $6 \times 10^{5}$ steps, which took 7 hours. Therefore, we can safely declare that using FishGym to train agile end-to-end swimming policies in an acceptable amount of time is completely feasible.

% Compared to \cite{zhang2022simulation}, our simulator achieves approximately 10 times faster than the traditional CFD simulators.

\begin{figure*}[t]
   \centering
   \includegraphics[width=0.99\linewidth]{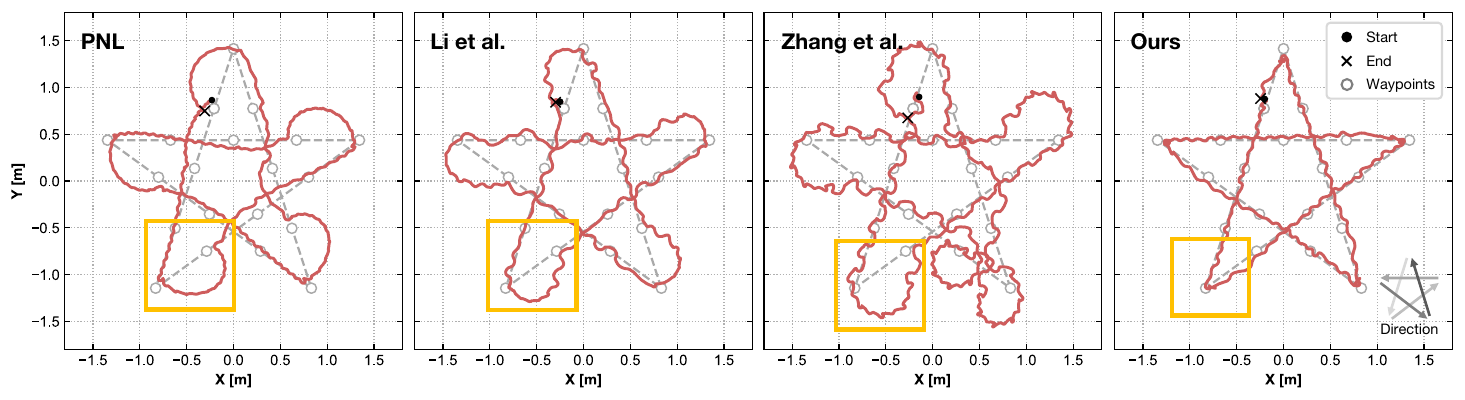}
   \vspace{-10pt}
   \caption{Multi-waypoints control task of the robotic fish. Experimental results of the robotic fish in real-world under four different controller to perform pentagram-shaped waypoints control task. When the robotic fish reaches the current waypoints, the next waypoints location will be provided. The yellow box emphasizes the performance of the robotic fish at the vertex turn.}
   \label{fig:waypoints}
\end{figure*}

\begin{figure}[t]
    \centering
    \includegraphics[width=0.95\linewidth]{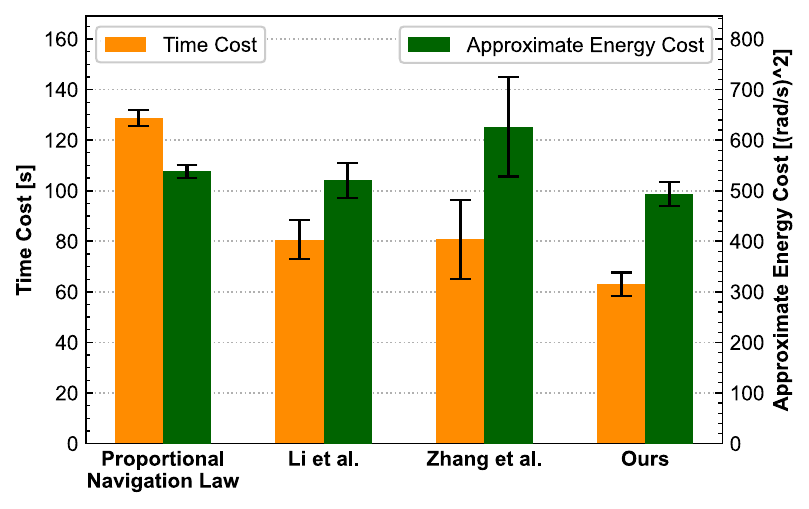}
    \vspace{-10pt}
    \caption{Statistical results of waypoints tracking tasks. Each controller was repeated three times.}
    \label{fig:statistic}
\end{figure}

\subsection{Sim-to-real Directly Transfer}\label{sec:transfer}
Next, the trained end-to-end swimming policy is directly transferred to the real-world without any fine-tuning. We performed position control tasks under the same condition in both simulation and real-world. Specially, the target position is placed \SI{1.5}{m} behind the robotic fish, which required the robotic fish have to make extremely sharp 180 degree turn to complete the task. The screenshot sequences of experiments in simulation and real-world are shown in \cref{fig:snapshot}. We can clearly see that the entire completion process consists of three stages: Firstly, the robotic fish bends and contracts violently to quickly adjust its orientation towards the target position with only \SI{1}{s}. Then, the robotic fish performs high-frequency rapid oscillations to accelerate the velocity as fast as possible. Finally, the robotic fish stop swinging and use the speed momentum to approach the target to precisely adjust the bending degree of the body to control the direction, efficiently and agilely reaching the target with totally \SI{5}{s}. This multi-modal motion ability is impressive, especially considering that this policy is purely trained in CFD simulation and directly transferred to real-world without any fine-tuning.

By comparing the screenshots of simulation and reality in \cref{fig:snapshot}, we can draw another important conclusion, that is, thanks to the sim-to-real techniques presented in \cref{sec:method}, even under agile movements with large swings, the policy trained using proposed framework still ensure highly consistency between simulation and reality. This makes the framework of training, debugging, and evaluating policy using pure CFD simulation more promising and attractive. We strongly recommend readers to watch the accompanying video to intuitively experience the agility of the trained policy and the consistency between simulation and reality.

\subsection{Pentagram Waypoints Task}
We further apply the trained position control policy to multi-waypoints task in real-world for comprehensive evaluation. The multi-waypoints task can be seen as a concatenation of multiple position control tasks, and when the robotic fish reaches the current waypoint, the next waypoint will be assigned. Specifically, there are a total of 21 waypoints arranged in the challenging shape of a pentagram, which requires the controller to be agile enough to achieve a large angle sharp turn of nearly 145 degree at the vertex during the motion process.

We also considered three state-of-the-art robotic fish control methods for comparison, namely: 1) \emph{Proportional Navigation Law (PNL)} \cite{guelman1971qualitative}, which is commonly used for tasks such as point-to-point tracking of robotic fish, 2) \emph{Li et al.} \cite{li2019bottom}, which combines Active Disturbance Rejection Control (ADRC) \cite{han2009pid} with PNL to improve the motion performance and disturbance resistance of robotic fish, and 3) \emph{Zhang et al.} \cite{zhang2022simulation}, which is the latest control framework based on RL that can generate various control policies for fish-like robots.

The experimental trajectory results of each controller are shown in \cref{fig:waypoints}. We can see that the \emph{PNL} has a large turning radius at the vertex and requires a certain distance to restore linear motion after making large angle turns. \emph{Li et al.} requires a smaller radius for turning, but has obvious fluctuations in straight lines. \emph{Zhang et al.} was more likely to miss waypoints due to its discrete action space, resulting in more additional consumption. In contrast, the proposed \emph{Ours} not only achieves stable periodic oscillation in straight lines, but also achieves large angle turning performance through non-periodic oscillation at the vertices, thereby obtaining a accurate pentagram trajectory.

We conducted three repeated experiments for each controller and organized the statistical results as shown in \cref{fig:statistic}. We consider two indicators: spending time and energy losses. The calculation formula for energy loss is $\hat{e} = \sum_{t=0}^{T}||\dot{\mathbf{J}}_{t}||^{2}$, which approximates energy using the integral of joint angular velocity at each timestamp. The statistical results indicate that the proposed \emph{Ours} has significantly shorter time consumption under similar energy consumption.

\subsection{Ablation of Sim-to-real Techniques}
In order to further analyze the role of the proposed sim-to-real techniques in policy transfer, we conducted ablation experiments on the three techniques used in this work to quantitatively demonstrate their impacts. Specifically, we compare the statistical results of the average reward of total four control groups under 180 degree turn tasks introduced in \cref{sec:transfer} and repeated it by five times. Among them, those parameters that do not use technology are configured by default according to the default parameters of the FishGym paper \cite{liu2022fishgym}.

\begin{figure}[t]
    \centering
    \includegraphics[width=0.94\linewidth]{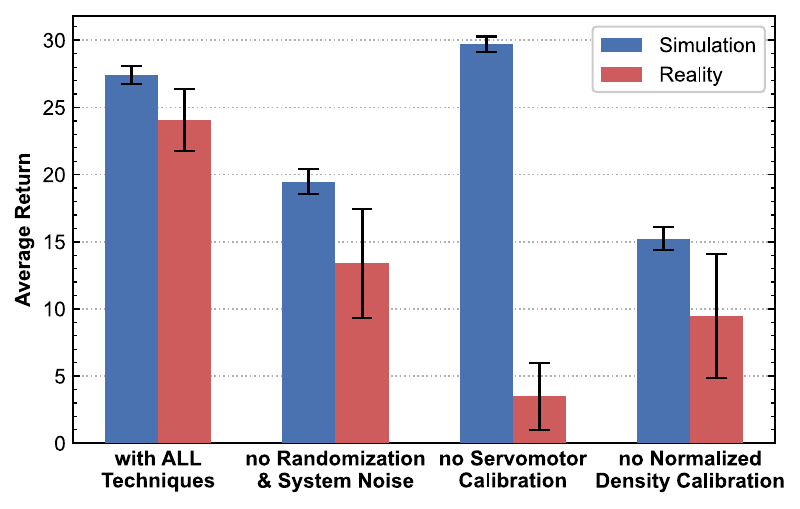}
    \vspace{-10pt}
    \caption{Performance of policies trained using different sim-to-real techniques in the simulation and real-world for the 180 degree position control. Policies are trained using only data from simulation and directly transfer to the real-world without fine-tuning.}
    \label{fig:ablation}
\end{figure}

As statistical results shown in \cref{fig:ablation}, \emph{with all techniques} has the best performance in reality, and the absence of any technique leads to a significant decrease in real-world performance. The most significant performance decline is \emph{no servomotor calibration}, that this means that actuator calibration plays an important role in RL policy sim-to-real transfer, as demonstrated in \cite{hwangbo2019learning}.

In addition, there is still a discrepancy between the simulation and real-world even using all techniques. We believe that is caused by two factors: the first factor is that there is a difference in the rubber appearance of the simulated and real robotic fish, which leads to different dynamic effects during the interaction with hydrodynamics. The second factor is that the surface of the real robotic fish is composed of multiple materials (ABS plastic material for the head and tail, rubber material for the body), which results in different surface viscous forces between the simulation and reality, affecting the dynamics of the robotic fish.

%%%%%%%%%%%%%%%%%%%%%%%%%%%%%%%%%%%%%%%%%%%%%%%%%%%%%%%%%%%%%%%%%%%%%%%%%%%%%%%%
\section{Conclusion}\label{sec:conclusion}

In this letter, we introduce a novel sim-to-real framework that learns a zero-shot transferable strategies from an advanced high-performance CFD simulator. Leveraging RL model-free and end-to-end nature, this framework eliminates the need for expert-derived dynamic models (simplifying its use) and enhances the robotic fish's ability and energy efficiency. Using position control as an example, we demonstrate the framework's high consistency between simulation and physical results and superiority over the classic solution.

While the proposed approach enhances the robotic fish’s mobility, it still requires learning to swim from scratch for new tasks. Future work will focus on accelerating training by avoiding redundant learning and targeting specific tasks, thus improving the efficiency of the control strategy development.

%%%%%%%%%%%%%%%%%%%%%%%%%%%%%%%%%%%%%%%%%%%%%%%%%%%%%%%%%%%%%%%%%%%%%%%%%%%%%%%%%%%%%%%%%%%%%%%%%%%%
% \section*{Acknowledgments}\label{sec:acknowledgment}

%%%%%%%%%%%%%%%%%%%%%%%%%%%%%%%%%%%%%%%%%%%%%%%%%%%%%%%%%%%%%%%%%%%%%%%%%%%%%%%%
% \addtolength{\textheight}{-10cm}  % This command serves to balance the column lengths

%%%%%%%%%%%%%%%%%%%%%%%%%%%%%%%%%%%%%%%%%%%%%%%%%%%%%%%%%%%%%%%%%%%%%%%%%%%%%%%%
\bibliographystyle{IEEEtran}
\bibliography{refs}

\end{document}